\documentclass[letterpaper, 10 pt, journal, twoside]{IEEEtran}
\usepackage{cite}

\ifCLASSINFOpdf
  \usepackage[pdftex]{graphicx}
\else
\fi
\usepackage{amsmath}

\usepackage{multirow}

\begin{document}
%
\title{GNSS Odometry: Precise Trajectory Estimation Based on Carrier Phase Cycle Slip Estimation}
%
%
%

\author{Taro Suzuki$^{1}$%
\thanks{Manuscript received: February 24, 2022; Revised May 16, 2022; Accepted June 6, 2022.}
\thanks{This letter was recommended for publication by Associate Editor F. Caballero and Editor J. Civera upon evaluation of the reviewers’ comments.}
\thanks{This work was supported by JSPS KAKENHI under Grant 19H00750.}
\thanks{The author is with the Future Robotics Technology Center, Chiba Institute of Technology, Chiba 275-0016, Japan (e-mail: taro@furo.org).}%
\thanks{Digital Object Identifier 10.1109/LRA.2022.3182795}
}
%
%

\markboth{IEEE ROBOTICS AND AUTOMATION LETTERS. PREPRINT VERSION. ACCEPTED June, 2022}
{Taro Suzuki: GNSS Odometry: Precise Trajectory Estimation Based on Carrier Phase Cycle Slip Estimation} 

%



\maketitle

\begin{abstract}
  This paper proposes a highly accurate trajectory estimation method for outdoor mobile robots using global navigation satellite system (GNSS) time differences of carrier phase (TDCP) measurements. By using GNSS TDCP, the relative 3D position can be estimated with millimeter precision. However, when a phenomenon called cycle slip occurs, wherein the carrier phase measurement jumps and becomes discontinuous, it is impossible to accurately estimate the relative position using TDCP. Although previous studies have eliminated the effect of cycle slip using a robust optimization technique, it was difficult to completely eliminate the effect of outliers. In this paper, we propose a method to detect GNSS carrier phase cycle slip, estimate the amount of cycle slip, and modify the observed TDCP to calculate the relative position using the factor graph optimization framework. The estimated relative position acts as a loop closure in graph optimization and contributes to the reduction in the integration error of the relative position. Experiments with an unmanned aerial vehicle showed that by modifying the cycle slip using the proposed method, the vehicle trajectory could be estimated with an accuracy of 5–30 cm using only a single GNSS receiver, without using any other external data or sensors.
\end{abstract}

\begin{IEEEkeywords}
Localization, SLAM, Odometry, GNSS
\end{IEEEkeywords}

%
\IEEEpeerreviewmaketitle

%
%
%
%

\section{Introduction}
\IEEEPARstart{T}{he} accurate self-positioning of outdoor robots is an essential fundamental technology in various fields, such as the automatic operation of unmanned robotic systems, mapping, and delivery services. A global navigation satellite system (GNSS) can be used for outdoor position estimation. It allows the estimation of the absolute position in a coordinate system fixed to the Earth. In particular, the real-time kinematic (RTK)-GNSS technique uses an additional GNSS reference station, and the double difference between receiver-to-receiver and satellite-to-satellite GNSS observations, such as pseudorange and carrier phase measurements, cancels most of the errors in GNSS observations and enables position estimation with centimeter accuracy \cite{GNSS_General}. However, RTK-GNSS requires the installation of GNSS reference stations within 10–20 km of the usage environment, and network RTK-GNSS, which utilizes the existing receiver networks, also requires the communication of GNSS correction data. In recent years, the use of GNSS precise point positioning has increased \cite{GNSS_Handbook}. This stand-alone positioning technique uses high-precision satellite orbit and clock information. However, its position convergence time is too long to be applicable to mobile robot positioning. Therefore, there is a need for a high-precision positioning method that uses only a GNSS receiver, without the need for external data or sensors.

In recent years, a method called time differences of carrier phases (TDCP), which uses the time difference of GNSS carrier phase measurements, has been actively studied as a high-precision relative position estimation method using a single GNSS receiver \cite{TDCPEKF,TDCEKF2}. The TDCP-based technique can estimate the relative position (velocity) with high accuracy by canceling the ionospheric delay, tropospheric delay, satellite clocks and orbit errors, etc. in GNSS observations, by considering the time difference of the GNSS carrier phase measurements. Although various TDCP algorithms have been studied, one of the major problems is the occurrence of a large relative position error when the GNSS carrier phase becomes discontinuous. This phenomenon is known as cycle slip. It occurs in the following environments: if GNSS signals are blocked or obtained from satellites with a low elevation angle, antenna motions have large accelerations, or if the carrier phase of the GNSS signal is out of track.


\begin{figure}[t]
   \centering
   \includegraphics[width=85mm]{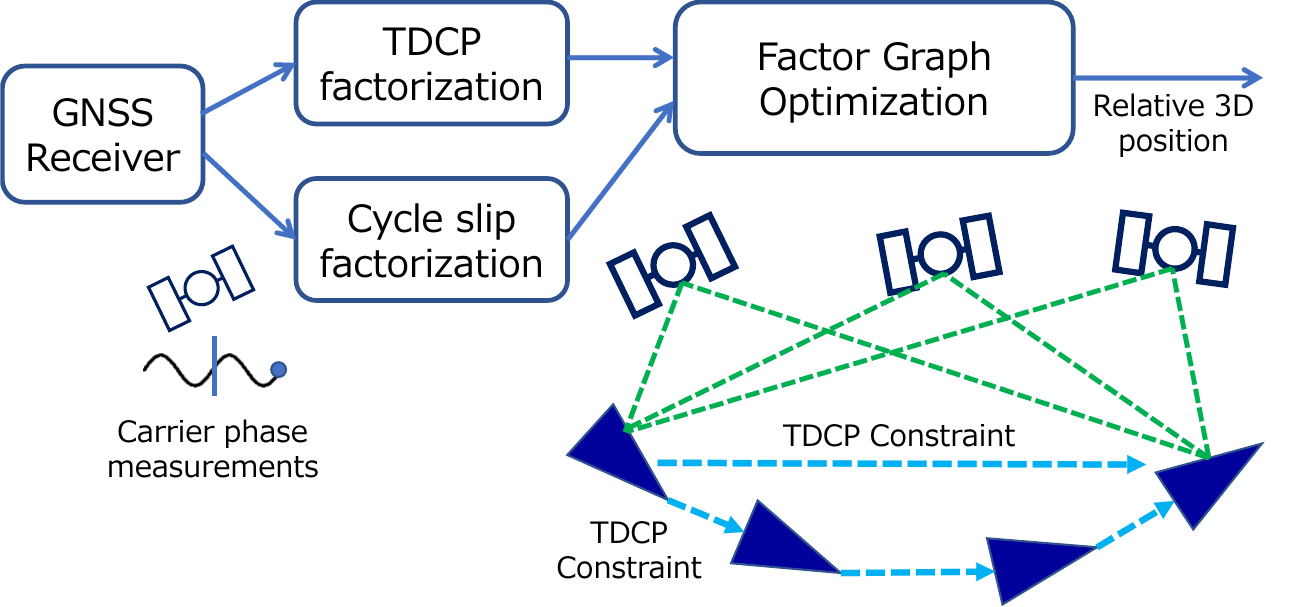} 
   \caption{Overview of the proposed method of GNSS odometry. This method uses only GNSS receivers to construct a graph structure modeled by the time difference of carrier phase (TDCP) measurements, and carrier phase cycle slips. Moreover, it enables loop closing in the graph optimization using only GNSS observations and thus enables a highly accurate trajectory estimation.}
   \label{fig1}
\end{figure}
In the field of robotics, simultaneous localization and mapping (SLAM), a robot position and pose estimation method based on graph optimization using camera and lidar observations, has been actively studied \cite{slam1,slam2,slam3}. In SLAM, to reduce the state-estimation error, the generation of loop-closure edges, which are constraints between nodes in a graph that are spatially and temporally separated, significantly contributes to the state-estimation accuracy. In our previous study, we had proposed a graph optimization method called time-relative (TR)-RTK-GNSS, which utilizes loop-closure edge generation and a single GNSS \cite{TRRTK}. This method uses the double difference of the TDCP measurements to solve the integer ambiguity in the carrier phase measurements; additionally, it utilizes a high-precision relative position constraint for graph optimization.

However, this study has some limitations as follows:
\begin{itemize}
   \item It is necessary to calculate the difference in TDCP between satellites (double difference), which degrades performance when the noise of the carrier phase measurement of the reference satellite is too large.
   \item The relative position constraint is added to the graph (loosely coupled integration) only when the carrier phase integer ambiguity is resolved. TR-RTK-GNSS requires at least five to six satellites for positioning calculation, and if the number of observed satellites is decreased, the constraint cannot be added to a graph, and the final trajectory estimation accuracy decreases.
\end{itemize}

This study is a further development and generalization of our previous work. Fig. 1 shows a schematic of the trajectory estimation method proposed in this study. The main changes in this method are as follows:
\begin{itemize}
   \item TDCP measurements were directly used as the constraint (tightly coupled integration) instead of the double difference of the carrier phase. The relative position constraint can be used even when only one satellite is present.
   \item Cycle slips in the carrier phase were added to the estimated state, and the amount of cycle slips was explicitly estimated. This makes it possible to use the observed values of low-elevation satellites with frequent cycle slips for obtaining relative position constraints.
\end{itemize}

\subsection{Related Studies}
Various sensor-based methods have been studied for trajectory estimation in mobile robots, including visual odometry using cameras, lidar odometry, and a combination of these methods with inertial measurement units (IMUs) \cite{VOsurvey,LOsurvey}. Trajectory estimation using cameras and lidar is difficult to apply in environments where observation information is sparse and there are few features or open environments such as the sky. In this study, we focused on trajectory estimation using the GNSS. GNSS point positioning accuracy, which uses a pseudorange, is at the meter level owing to the limitation of the observation accuracy of the pseudorange. As with odometry using cameras and lidars, a method was used to estimate the trajectory by integrating the velocity estimated using GNSS, which is commonly based on the Doppler shift (pseudorange rate) of GNSS signals \cite{vel1}. Doppler measurement does not include cycle slip or carrier phase ambiguity, and the relative velocity between the satellite and the receiving antenna can be measured directly. Similar to the conventional positioning using a pseudorange, the 3D velocity and receiver clock drift can be estimated from Doppler shift measurements from four or more satellites using the least-squares method.

On the other hand, the relative position and velocity estimation using TDCP has been studied \cite{td1,td2,td3,vel2,TDCPEKF,TDCEKF2}. The time-sequential carrier phase difference contains the relative distance between the satellite and receiving antenna with high accuracy if the carrier phase ambiguity is time-invariant. The velocity estimated using TDCP has been reported to be more accurate than that estimated using Doppler-based velocity estimation \cite{td1}. Reference \cite{td1} showed that the accuracy of the velocity estimation using TDCP in the static test was 2 to 3 mm/s. Because the estimation of the velocity/relative position using TDCP assumes continuity of the carrier phase, the relative position cannot be calculated correctly when cycle slip occurs, which breaks the continuous tracking of the carrier phase. In references \cite{TDCP_rubust} and \cite{gsdc}, to cope with this problem, the TDCP observations where cycle slips occur are excluded as outliers using a robust optimization method. In \cite{TDCP_rubust}, M-estimation was used to address cycle slips. Switchable constraints \cite{gognss0}, which simultaneously estimate the weight of each constraint, were used for GNSS positioning in \cite{gsdc}. Here, cycle slip can sometimes occur in minimal cycles, in which case, it is difficult to completely exclude its effect using the robust optimization technique. In addition, because cycle slips frequently occur for satellites with low elevation angles, the rejection of observations from low-elevation satellites deteriorates the satellite geometry, which, in turn, deteriorates the vertical positioning accuracy. 

Our previous work \cite{TRRTK}, have been conducted to estimate cycle slips using the carrier phase ambiguity resolution technique \cite{lambda}. This approach enables us to estimate the trajectory of an unmanned aerial vehicle (UAV) with an accuracy of 20–30 cm using only a low-cost single-frequency GNSS receiver. Because these methods calculate the difference in observations between satellites (double difference) in addition to the time difference, the problem of selecting a reference satellite from among multiple GNSS systems arises. In addition, this method cannot be used when the number of satellites is decreased. On the other hand, in this study, we propose a tightly coupled graph optimization method using TDCP as the direct observation without double difference. Moreover, the cycle slip of each carrier phase is added to the estimated state.

\subsection{Contributions}
The contributions of this study are as follows. 
\begin{itemize}
\item We proposed a tightly coupled integration with TDCP observations to estimate the vehicle trajectory. In addition, we considered cycle slips, which were incorporated into the state estimation.
\item We constructed an observation model of TDCP that includes cycle slip in factor graph optimization.
\item The vertical error of the GNSS was reduced by using the TDCP of low-elevation-angle satellites.
\end{itemize}

\section{TDCP Observation Model}
\subsection{Carrier Phase Observation Model}
The observation model for the GNSS carrier phase of satellite $k$ at time $t$ is given by the following equation:

\begin{equation}
   \lambda \Phi_{t}^{k}=r_{t}^{k}+\lambda B_{t}^{k}+c\left(\delta t_{t}-\delta T_{t}^{k}\right)+D_{t}^{k}-I_{t}^{k}+T_{t}^{k}+\epsilon_{t}^{k}
\end{equation}

where $\lambda$ is the signal wavelength, $\Phi_{t}^{k}$ is the measured carrier phase in cycles, $r_{t}^{k}$ is the satellite-receiver geometric distance, $B_{t}^{k}$ is the carrier phase integer ambiguity, $c$ is the speed of light. $\delta t_{t}$ and $\delta T_{t}^{k}$ are the receiver and satellite clock biases, respectively. $D_{t}^{k}$, $I_{t}^{k}$, and $T_{t}^{k}$ are the satellite orbit, ionosphere, and troposphere errors, respectively, and $\epsilon_{t}^{k}$ is the carrier phase measurement noise, which includes phase wind-up effect and carrier phase multipath error. The time difference between the carrier phase measurements $\Delta \Phi_{t_2,t_1}^{k}$ at times $t_1$ and $t_2$ was calculated as follows:

\begin{IEEEeqnarray}{lCr}
   \lambda \Delta \Phi_{t_2,t_1}^{k} = \lambda \left[\Phi_{t_2}^{k}-\Phi_{t_1}^{k} \right] \nonumber\\
   = \Delta r_{t_2,t_1}^{k} + \lambda \left( B_{t_2}^{k}-B_{t_1}^{k} \right) + c \Delta\delta t_{t_2,t_1}-c\Delta \delta T_{t_2,t_1}\nonumber\\
   +\Delta D_{t_2,t_1}-\Delta I_{t_2,t_1}+\Delta T_{t_2,t_1}+\Delta \epsilon_{t_2,t_1}
\end{IEEEeqnarray}

where, $\Delta$ denotes the differencing operation. In a normal TDCP, the time difference is calculated between consecutive epochs of less than one second. In this case, the time variation of each error component, except the receiver clock change $\Delta\delta t_{t_2,t_1}$ can be ignored, and the exact distance change $\Delta r_{t_2,t_1}^{k}$ between the satellite and the receiver can be calculated. However, in this study, the time difference $t_2-t_1$ is calculated at a maximum of 60 s because it is used as a loop-closure for graph optimization. Therefore, the difference in each error component is not negligible. Each of these error components is treated as follows:

\subsubsection{Ionospheric Delay and Tropospheric Delay}
The time difference of ionospheric and tropospheric delays $\Delta I_{t_2,t_1}$ and $\Delta T_{t_2,t_1}$ include temporal variations in the delay itself and variations due to changes in the propagation path between the satellite and the receiver. In \cite{td1}, it was shown that these values could be ignored when generating TDCP observations. It is almost negligible if the time difference is less than a few seconds. However, if the difference is not between consecutive GNSS observations, but between observations several tens of seconds apart, it may become non-negligible, particularly for low-elevation satellites. When dual-frequency GNSS observations are available, the ionospheric delay error can be canceled using the ionosphere-free linear combination. However, the coupling of carrier phases at multiple frequencies increases the noise, and the cycle slip is not an integer. Therefore, to reduce the difference between the ionosphere and troposphere errors, the ionospheric and tropospheric delays were corrected by the model at the time of each observation before the carrier phase difference was calculated. The ionospheric delay was corrected using the Klobuchar model \cite{GNSS_General} and the tropospheric delay was corrected using the Saastamoinen model \cite{GNSS_General}.

\subsubsection{Satellite Clocks and Orbit Errors}
Among the satellite clocks and orbit errors, the satellite clock error is the most dominant error in TDCP observations \cite{TDCP2}. When the time difference of the carrier phase observation becomes large, the relative position estimation accuracy of TDCP deteriorates owing to the variation in the satellite clock error. To reduce the error caused by the time variation of the satellite clock error, it is possible to use a precise ephemeris, which provides highly accurate satellite orbit and clock information. However, using the precise ephemeris in real time requires external communication; therefore, in this study, the satellite clock and orbit error was corrected using the ordinary broadcast ephemeris.

The TDCP measurement between times $t_1$ and $t_2$, corrected for the error described above $\Delta \widehat{\Phi}_{t_2,t_1}^{k}$ is expressed as follows:

\begin{IEEEeqnarray}{lCr}
   \lambda \Delta \widehat{\Phi}_{t_2,t_1}^{k} \nonumber \\
   = \lambda \Delta \Phi_{t_2,t_1}^{k} - \Delta D_{t_2,t_1} + \Delta I_{t_2,t_1} - \Delta T_{t_2,t_1} + c\Delta \delta T_{t_2,t_1} \nonumber\\
   \simeq \Delta r_{t_2,t_1}^{k} + c \Delta\delta t_{t_2,t_1} + \lambda \left( B_{t_2}^{k}-B_{t_1}^{k} \right) +\Delta \epsilon_{t_2,t_1}
\end{IEEEeqnarray}

The corrected TDCP measurement retains the change in the satellite-receiver distance, the change in the carrier phase ambiguity (cycle slip) between times $t_1$ and $t_2$, and the receiver clock change.

\begin{figure*}[t]
   \centering
   \includegraphics[width=155mm]{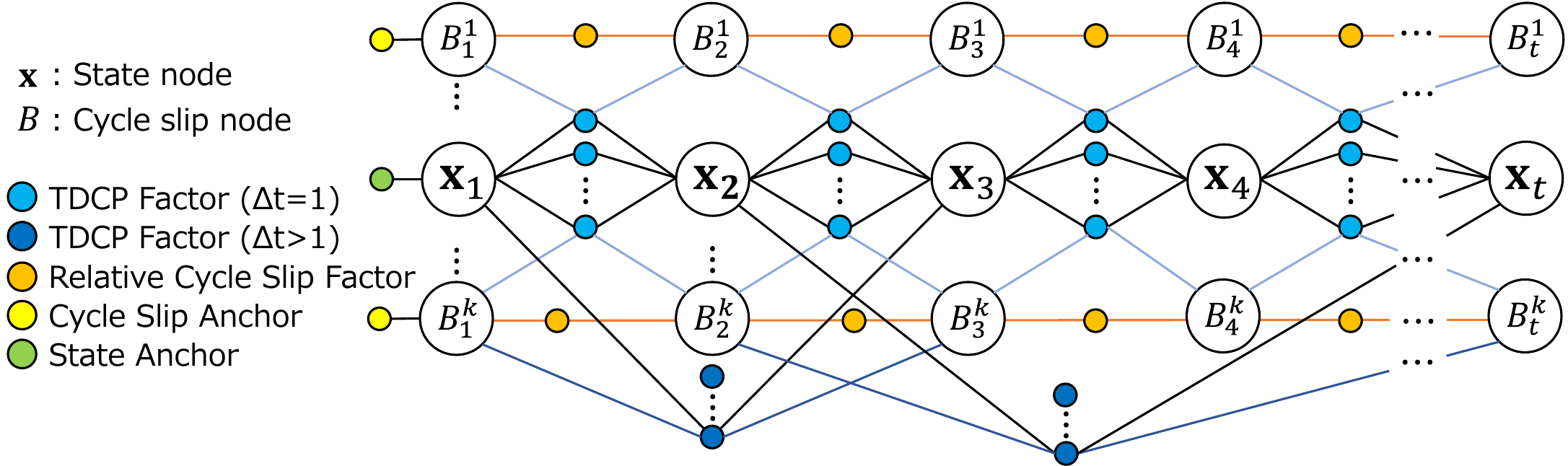} 
   \caption{The proposed factor graph model. The relative 3D position constraint between nodes separated by the TDCP observations from each satellite at times $t_1$ and $t_2$ (TDCP factor). The accumulated cycle slip amount in the carrier phase measurements of each satellite is added to the state, and the constraint of time variation of cycle slip is added between successive cycle slips (relative cycle slip factor).}
   \label{fig2}
\end{figure*}

\subsubsection{Relative Position Estimation}
The 3D position at time $t$ is defined in the earth-centred earth-fixed (ECEF) coordinate system as follows:

\begin{equation}
   \mathbf{p}_{t}=\left[\begin{array}{lll}
   x_t & y_t & z_t
   \end{array}\right]^{T}
\end{equation}

The change in the distance between satellite $k$ and antenna position $\Delta r_{t_2,t_1}^{k}$ at times $t_1$ and $t_2$ can be expressed as follows:

\begin{equation}
   \Delta r_{t_2,t_1}^{k} = \Delta L_{t_2,t_1}^{k}-\mathbf{u}_{t_2}^{k} \cdot \Delta \mathbf{p}_{t_2,t_1}
\end{equation}

Here, $\Delta L_{t_2,t_1}$ is the change in distance owing to satellite motion from the antenna position. $\mathbf{u}_{t_2}^{k}$ is the line-of-sight vector in the ECEF coordinate system from the receiver's approximate position to satellite $k$, and $\Delta \mathbf{p}_{t_2,t_1}$ is the relative antenna 3D position between times $t_1$ and $t_2$. The observation equation for TDCP is obtained as follows:

\begin{IEEEeqnarray}{lCr}
   \lambda \Delta \widehat{\Phi}_{t_2,t_1}^{k} \nonumber - \Delta L_{t_2,t_1}^{k} \nonumber\\
   =  -\mathbf{u}_{t_2}^{k} \cdot \Delta \mathbf{p}_{t_2,t_1} + c \Delta\delta t_{t_2,t_1}+ \lambda \left( B_{t_2}^{k}-B_{t_1}^{k} \right) +\Delta \epsilon_{t_2,t_1}
\end{IEEEeqnarray}

If there is no cycle slip, the carrier phase ambiguity is $B_{t_2}^{k}-B_{t_1}^{k}=0$ and can be eliminated. Using the least-squares method, the relative 3D position $\Delta \mathbf{p}_{t_2,t_1}$ and clock change $\Delta\delta t_{t_2,t_1}$ between times $t_1$ and $t_2$ can be estimated. The measurement accuracy of the carrier phase was approximately 3 mm in terms of standard deviation, and the accuracy of the relative position was also accurate at a millimeter level. However, when a cycle slip of one cycle occurs, the observed distance between the satellite and the antenna jumps by approximately 0.2 m because the wavelength of GNSS L1 signal is approximately 0.2 m. This has a significant impact on the precise relative position estimation.

\section{Graph Structure}
Factor graph optimization has been used in recent years in robotics to solve various time-series state-estimation problems \cite{go1}. A factor graph has variable nodes $\mathbf{X}$ and factor nodes that are connected by edges. The edges between the factor and variable nodes encode the error functions, $\mathbf{e}(\cdot)$. Various edges (constraints) can be used for state-estimation problems. In recent years, graph optimization has been actively studied in the field of GNSS \cite{gognss1,gognss2,gognss3,gognss4}.

The factor graph structure used in this study is illustrated in Fig. 2. As a node of the graph, we used the 3D position in the ECEF coordinate system at each time and the receiver clock bias. The set of state nodes is represented as follows.

\begin{equation}
   \mathbf{X}=\left[\begin{array}{llll}
      \mathbf{x}_1 & \mathbf{x}_2 & \cdots & \mathbf{x}_n
   \end{array}\right]
\end{equation}

The state at time $t$ is as follows:

\begin{equation}
   \mathbf{x}_{t}=\left[\begin{array}{ll}
      \mathbf{p}_{t} & \mathbf{t}_{t}
   \end{array}\right]^{T}
\end{equation}

where, $\mathbf{t}_{t}$ consists of the clock bias of each satellite system.

\begin{equation}
   \mathbf{t}_{t}=\left[\begin{array}{llll}
   {t_{\mathrm{GPS}, t}} & {t_{\mathrm{GLO}, t}} & {t_{\mathrm{GAL}, t}} & {t_{\mathrm{BDS}, t}}
   \end{array}\right]^{T}
\end{equation}

Note that the clock bias of GLONASS, Galileo, and BeiDou is estimated as the bias with respect to GPS time. Here, a node for the cumulative cycle slip value from the initial state,  $B_{t}^k$ is added to the carrier phase observation of each satellite, where the unit of $B_{t}^k$ is the number of cycles, which is an integer value. The relative cycle slip factor (orange circle in Fig. 2) was added between the cycle slip nodes as a constraint on the amount of relative cycle slip. Note that, in Fig. 2, the number of satellites is kept constant for clarity. When a satellite becomes invisible, the cycle slip node $B_{t}^k$ and relative cycle slip factor are added to the graph, but the TDCP factor is not.

As mentioned above, the time difference information of the carrier phase is included in the relative position change between the nodes, and the absolute position cannot be estimated. The initial state node is constrained to the position of the ECEF coordinate system estimated by ordinary single-point positioning, using the GNSS pseudorange as an initial constraint to anchor the 3D position (light green circle in Fig. 2). Similarly, the initial cumulative cycle slip node is constrained to zero cycles  (yellow circle in Fig. 2).

\subsection{TDCP Factor with Cycle Slip}
Considering cycle slip, the error function of the TDCP factor can be defined as follows:

\begin{IEEEeqnarray}{lCr}
   e_{\mathrm{TD},t_2,t_1}^{k}=\mathbf{H}_{t_2,t_1}^{k} \left(\mathbf{x}_{t_2}-\mathbf{x}_{t_1}\right) \nonumber\\
   - \left\{\left( \lambda \Delta \widehat{\Phi}_{t_2,t_1}^{k} - \Delta L_{t_2,t_1}^{k} \right) - \lambda \left( B_{t_2}^{k}-B_{t_1}^{k} \right) \right\}
\end{IEEEeqnarray}

Here, measurement matrix $\mathbf{H}_{t_2,t_1}^{k}$ can be formulated as:

\begin{equation}
   \mathbf{H}_{t_2,t_1}^{k}=\left[\begin{array}{lllll}
      \mathbf{u}_{t_2}^{k,T} & {1} & {\delta_{\mathrm{GLO}}^{k}} & {\delta_{\mathrm{GAL}}^{k}} & {\delta_{\mathrm{BDS}}^{k}}
   \end{array}\right]
\end{equation}

where, $\mathbf{u}_{t_2}^{k,T}$ is the line-of-sight vector from receiver to satellite $k$. Moreover, $\delta_{\mathrm{GLO}}^{k}$, $\delta_{\mathrm{GAL}}^{k}$, or $\delta_{\mathrm{BDS}}^{k}$ is equal to 1 when the $k$-th GNSS satellite is GLONASS, Galileo, or BeiDou, respectively. The minimized error of the TDCP factor was calculated as follows:

\begin{equation}
   \left\|e_{\mathrm{TD}, t_2,t_1}^{k}\right\|_{\Omega_{\mathrm{TD}}^{k}}=e_{\mathrm{TD}, t_2,t_1}^{k} \hspace{1pt} \Omega_{\mathrm{TD}}^{k} \hspace{1pt} e_{\mathrm{TD}, t_2,t_1}^{k}
\end{equation}

where, $\Omega_{\mathrm{TD}}^{k}$ represents the information matrix of TDCP computed from the variance of the carrier phase measurement. The GNSS carrier phase variance is determined as a function of the satellite elevation angle. For the selection of $t_1$ and $t_2$, the TDCP factor was added as a loop-closure edge for up to 60 s, in addition to the time between the neighboring nodes ($t_2-t_1=1$ s in this study).

\subsection{Relative Cycle Slip Factor}
When there is no cycle slip, the amount of cycle slip $B_t^k$ is constant between successive epochs. The error function for the relative constraint between the cycle slips was set as follows:

\begin{equation}
   e_{\mathrm{B},t}^{k}= B_{t+1}^{k}-B_{t}^{k}
\end{equation}

The minimized error of the relative cycle slip factor is calculated as:

\begin{equation}
   \left\|e_{\mathrm{B}, t}^{k}\right\|_{\Omega_{\mathrm{B}}}=e_{\mathrm{B}, t}^{k} \hspace{1pt} \Omega_{\mathrm{B}} \hspace{1pt} e_{\mathrm{B}, t}^{k}
\end{equation}

The GNSS receiver generally outputs a cycle slip detection flag or a loop lock indicator (LLI) for the observed carrier phase. In the presence of an LLI flag, the cycle slip continuity constraint is set to low because the possibility of a cycle slip is high. When multiple frequency observations are available, cycle slip detection using a geometry-free linear combination is possible. Therefore, the cycle slip continuity constraints are set as low also when geometry-free linear combination jumps are detected \cite{gnss_understanding}. In this study, we set up two information matrices $\Omega_{\mathrm{B}}$, one for the constraint that determines the continuity of cycle slips under normal conditions, and the other for the likely cause of cycle slips and the switch between them. The values of these variances were determined heuristically based on experiments.

\subsection{Optimization and Implementation}
Finally, in the graph optimization, the state is estimated by minimizing the following equation.

\begin{IEEEeqnarray}{lCr}
   \widehat{\mathbf{X}}=\underset{\mathbf{X}}{\operatorname{argmin}} \sum_{t} \sum_{k}\left\|e_{\mathrm{TD}, t+1, t}^{k}\right\|_{\Omega_{\mathrm{TD}}^{k}}^{2}\nonumber\\
   +\sum_{t} \sum_{k}\left\|e_{\mathrm{TD}, t+\Delta t, t}^{k}\right\|_{\Omega_{\mathrm{TD}}^{k}}^{2} +\sum_{t}\sum_{k}\left\|e_{\mathrm{B},t}^{k}\right\|_{\Omega_{\mathrm{B}}}^{2} \nonumber\\
   +\sum_{k}\left\|e_{\mathrm{B},0}^{k}\right\|_{\Omega_{\mathrm{B},0}}^{2}+\left\|e_{\mathbf{x},0}\right\|_{\Omega_{\mathbf{x},0}}^{2}
\end{IEEEeqnarray}

Here, the last two terms represent the error function by the anchor factor to fix the initial values of the 3D position node $e_{\mathbf{x},0}$ and cycle slip node $e_{\mathrm{B},0}^{k}$. In particular, the initial 3D position of the state node is set to the ECEF position estimated by the single-point positioning, and the initial values of all cycle slip nodes are set to 0. The precise 3D relative position and amount of cycle slip can be estimated by minimizing this entire evaluation function.

For the actual implementation of the algorithm, we used GTSAM \cite{gtsam}, a generic open-source library for graph optimization, and the Gauss–Newton optimization solver. To handle GNSS observation data and estimate satellite position, clock, and the ionospheric and tropospheric delays, we used RTKLIB \cite{rtklib}, an open-source library widely used for GNSS analysis processing.

\begin{figure}[t]
   \centering
   \includegraphics[width=75mm]{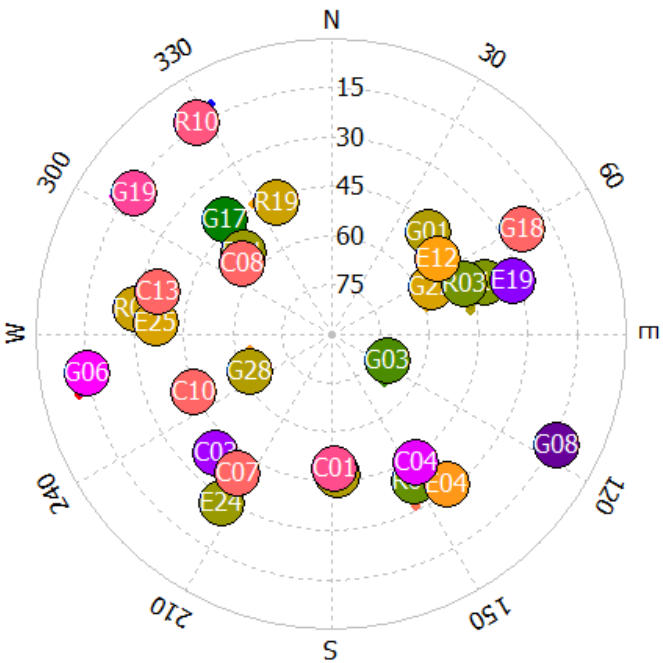} 
   \caption{GNSS satellite constellation during the static experiment. “G,” “J,” “C,” “E,” and “R” denote the  GPS, QZSS, BeiDou, Galileo, and GLONASS satellites, respectively.}
   \label{fig3}
\end{figure}
\begin{figure}[t]
   \centering
   \includegraphics[width=85mm]{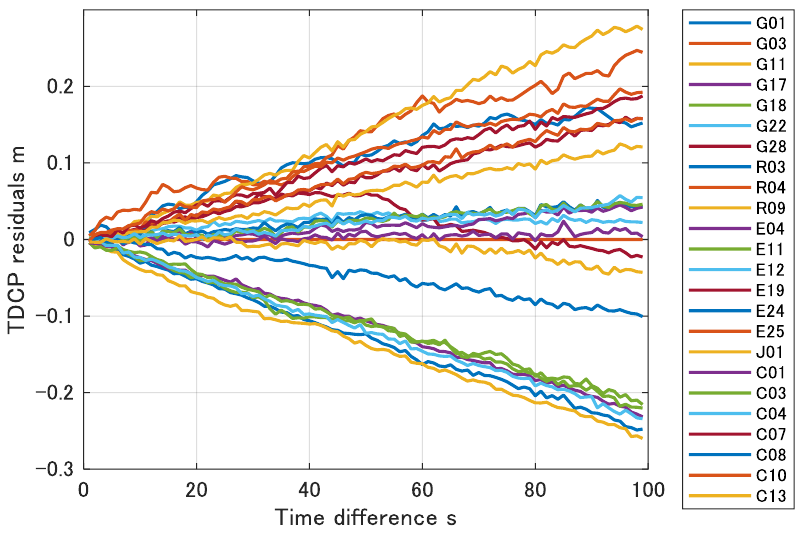} 
   \caption{Relationship between TDCP residuals and differential time using the proposed TDCP model. To cancel the receiver clock error, the TDCP difference between satellites is calculated using "G03" as the reference satellite.}
   \label{fig4}
\end{figure}

\section{Experiments}
\subsection{Static Test}
To evaluate the performance of the proposed method, we first evaluated the proposed TDCP residuals with respect to time difference in a static test. A GNSS receiver (Trimble BD930) was used in this test, and GNSS data were acquired at 1 Hz in an open-sky environment. The GNSS satellite constellation during the static test is illustrated in Fig.3. “G,” “J,” “C,” “E,” and “R” in Fig. 3 indicate GPS, QZSS, BeiDou, Galileo, and GLONASS satellites, respectively. The TDCP model, expressed as Eq. (3) includes the receiver clock error, and to evaluate the TDCP residuals, we calculated the difference in TDCP between satellites (double difference).

Fig. 4 shows the residuals of TDCP for each satellite. The "G03" satellite, which had the highest elevation angle, was selected as the reference satellite to calculate the double difference. From this figure, we can see that, as the time interval of TDCP increases, the residuals of TDCP also increase. This is mainly owing to the variation in the satellite clock error with time, as shown in Eq. (2). Although the satellite clock error was corrected using the broadcasting ephemeris, the TDCP error cannot be ignored when the time interval is large. However, if the time difference reaches 60 s, the relative position change can be measured with an accuracy of approximately $\pm$ 0.2 m. This study constructs a loop-closure edge based on the TDCP factor with an upper limit of 60 seconds. Note that the double difference of TDCP measurement includes cycle slips, but here we use a mask for low-elevation satellites and prior knowledge that the receiver is stationary to detect jumps in the TDCP, thus eliminating the effect of cycle slip, as shown in Fig. 4.

\begin{figure}[t]
   \centering
   \includegraphics[width=85mm]{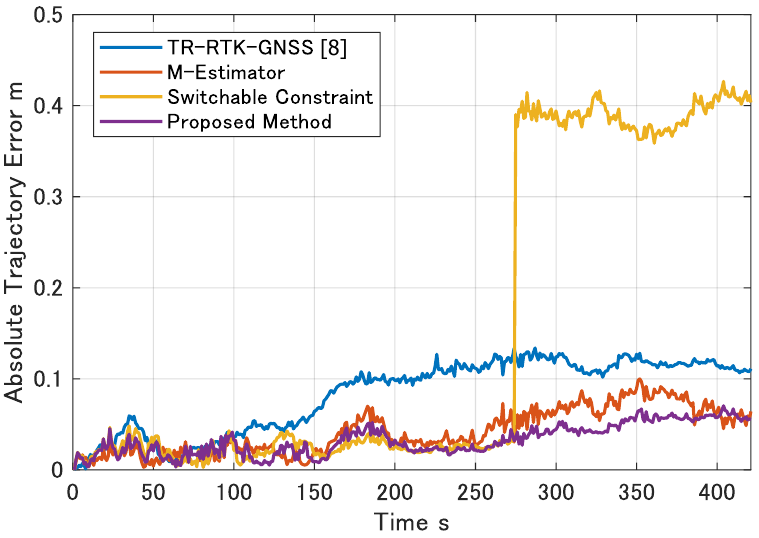} 
   \caption{Absolute trajectory error (ATE) of the proposed method and the methods considered for comparison in static test. As shown, the proposed method has the highest accuracy. In the method with robust optimization using switchable constraint (yellow line), a jump in the relative position was observed because of cycle slip.}
   \label{fig5}
\end{figure}

\begin{table}[]
   \centering
   \caption{Comparison of absolute trajectory error (ATE) of the proposed method and the existing methods in static test}
   \label{tab1}
   \begin{tabular}{ccc}
   \hline
   \multirow{2}{*}{Method}        & \multirow{2}{*}{RMS cm} & \multirow{2}{*}{MAX. cm} \\
                                  &                               &                              \\ \hline
   TR-RTK-GNSS \cite{TRRTK}              & 9.2                           & 13.39                        \\
   Robust optimization (M-Estimeter)           & 4.89                          & 9.99                         \\
   Robust optimization (Switchable Constraint) & 23.18                         & 42.67                        \\
   Ours (Cycle Slip Estimation)   & \textbf{3.68}       & \textbf{7.04}                         \\ \hline
   \end{tabular}
\end{table}

\begin{figure}[t]
   \centering
   \includegraphics[width=85mm]{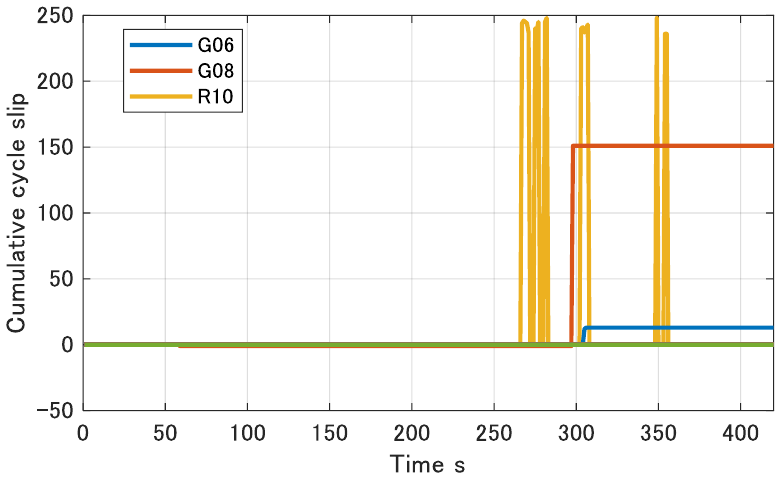} 
   \caption{Cumulative cycle slip estimated by the proposed method. Some cycle slips are observed for the satellites with low elevation angles.}
   \label{fig6}
\end{figure}

We evaluated the accuracy of the position estimation using static tests. As an evaluation metric, we used the absolute trajectory error (ATE), according to reference \cite{ATE}. The elevation mask of the satellite was set to 5$^\circ$ to use the observations of a satellite with low elevation angle.

To evaluate the proposed method, we compared it with the following methods. Each method was implemented using the same graph optimization framework, but with a modified observation model.

\begin{enumerate}
   \item Our previous paper using TR-RTK-GNSS \cite{TRRTK} 
   \item Robust optimization using M-estimator
   \item Robust optimization using switchable constraint \cite{gognss0} 
   \item Proposed method using cycle slip estimation
\end{enumerate}

In the implementation of M-estimator, Huber's influence function \cite{huber} was used, and the commonly used $c = 1.345$ was used as the tuning constant of the function. Fig. 5 and Table 1 show the ATE of each method in the static test. Here, the error plots continue to increase because of the accumulated relative position. Using the first method (TR-RTK-GNSS), the cumulative error increases slightly faster than in the case of the other methods. This is because TR-RTK-GNSS is loosely coupled, whereas other methods use the observations from each satellite independently by tightly coupled integration. The second and third methods can estimate the relative 3D position with good accuracy using robust optimization methods. However, the third method, which uses the switchable constraint, shows a large jump at approximately 270 s. This is probably because the effect of cycle slip of the carrier phase was not completely removed. Compared to these three methods, the proposed method, which uses cycle slip estimation, is the most accurate method for estimating the relative position.

Fig. 6 shows the cumulative cycle slip values estimated using the proposed method. In this figure, the carrier phase measurements from the low-elevation satellites “G06,” “G08,” and “R10”, as shown in Fig. 3, included several cycle slips. It was confirmed that the proposed method can estimate the cycle slip and maintain an accurate relative position by correcting the TDCP even after 270 s when the cycle slip occurs. The static test showed that the relative position estimation of the proposed method, which used only GNSS, was within approximately 5 cm over 400 s. 

\begin{figure*}[t]
   \centering
   \includegraphics[width=175mm]{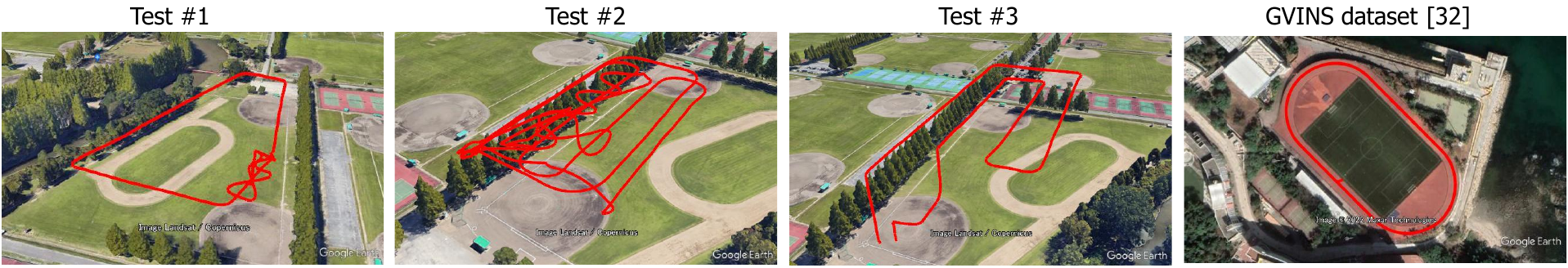} 
   \caption{Environment and trajectories. The three figures on the left show the UAV experiment, and the right shows the GVINS dataset \cite{GVINS}.}
   \label{fig7}
\end{figure*}
\begin{figure*}[t]
   \centering
   \includegraphics[width=175mm]{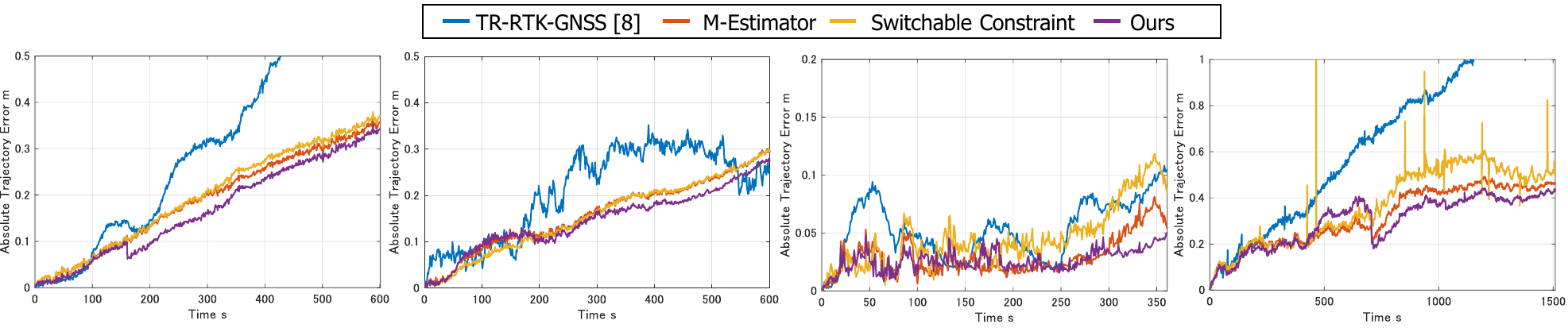} 
   \caption{Verification of the accuracy of the proposed and compared methods by kinematic tests. Similar to the static test, the proposed method was able to estimate the relative position with the highest accuracy in all tests.}
   \label{fig8}
\end{figure*}

\begin{table*}[]
   \centering
   \caption{Comparison of ATE of the proposed method and the existing methods}
   \label{tab2}
   \begin{tabular}{c|cc|cc|cc|cc}
   \hline
   \multirow{2}{*}{Method}      & \multicolumn{2}{c|}{Terst \#1}  & \multicolumn{2}{c|}{Test \#2} & \multicolumn{2}{c|}{Test \#3} & \multicolumn{2}{c}{GVINS dataset \cite{GVINS}} \\ \cline{2-9} 
                                    & RMS cm & MAX. cm & RMS cm & MAX. cm & RMS cm & MAX. cm & RMS cm & MAX. cm \\ \hline
   TR-RTK-GNSS \cite{TRRTK}                & 37.2         & 69.0        & 23.21        & 35.22       & 5.84         & 10.82       & 73.41        & 115.16       \\
   Robust optimization (M-Estimeter)             & 21.76        & 36.85       & 17.66        & 30.09       & 3.38         & 8.14        & 34.33        & 49.26        \\
   Robust optimization (Switchable   Constraint) & 22.74        & 37.98       & 17.48        & 30.22       & 5.28         & 11.83       & 40.81        & 255.85       \\
   Ours (Cycle Slip Estimation) & \textbf{19.72} & \textbf{35.23} & \textbf{16.12}     & \textbf{27.92}    & \textbf{2.81} & \textbf{5.06} & \textbf{31.89}       & \textbf{44.18}      \\ \hline
   \end{tabular}
\end{table*}

\subsection{Kinematic Test}
Kinematic positioning tests were conducted using an UAV. Currently, UAVs are widely used for various applications, including 3D mapping. More efficient 3D mapping can be achieved if only a single GNSS receiver can be used to accurately estimate the 3D trajectory of a UAV.

As in the static test, the UAV flight trajectory is evaluated using the proposed method. The UAV flight trajectories in the three experiments are shown in Fig. 7. Each flight lasted approximately 6–10 min, and had an altitude of 35 m and the UAV speed was varied between 2–3 m/s. We used the APX-15 UAV, which is a combination of a GNSS receiver and a high-grade inertial navigation system, as the ground truth of the UAV position. The positioning accuracy of APX-15 was 2 cm according to catalog specifications \cite{apx}. GNSS reference stations for the ground truth were set up in the vicinity of the experimental environment, but they were used neither for the proposed method not for those used for comparison. A single-frequency low-cost GNSS receiver (u-blox M8T) was used in the kinematic test. GNSS observations were acquired at 1 Hz, as in the static test, and the satellite elevation mask was set to 5$^\circ$.

The respective 3D ATEs for each test are shown in Fig. 8 and detailed in Table 2. Relative position estimation using TR-RTK-GNSS showed a faster increase in error with time than the other methods. However, the method using robust optimization can suppress the increase in accumulated error due to cycle slips. Furthermore, the proposed method that utilizes cycle slip estimation is the most accurate method for estimating the UAV trajectory. The data in Table 2 clarifies that, depending on the experimental environment and conditions, the cumulative error of the trajectory estimation of the proposed method was approximately 30 cm after approximately 10 min of flight.

\subsection{Kinematic Test using Open Dataset}
In addition to UAV experiments, the performance of the proposed method was evaluated using open datasets. Several datasets exist as benchmarks for the localization of mobile robots. However, the proposed method requires full GNSS raw measurements (e.g., pseudorange, carrier phase, and Doppler). Among the several datasets that provide raw GNSS data for localization benchmarking \cite{M2DGR,urbannav,urbanloco}, we used the GVINS dataset \cite{GVINS}. The sports field data in this dataset were acquired in an almost open-sky environment and included data acquired at 10 Hz using a u-blox F9P GNSS receiver, providing RTK-GNSS solutions as the ground truth for the entire data duration.

Fig. 7 shows the trajectory of the GVINS dataset, and the ATE of each method is plotted in Fig.8 and detailed in Table 2. As in the kinematic experiments, the proposed method was able to estimate trajectories more accurately than the other methods.

\subsection{Discussion}
The experimental results show that the proposed method can estimate the 3D relative position with a higher accuracy than the conventional trajectory estimation methods using TDCP. The following points are considered to contribute to this result:
\begin{itemize}
   \item The observed values of the carrier phase at each time point were used directly as constraints. This makes it possible to treat the observations from each satellite independently, which is not the case with the conventional TR-RTK-GNSS method. It is also expected that the constraint can be added, even when the number of satellites decreases.
   \item By explicitly estimating the cycle slip of the carrier phase, the observations of the carrier phase from low-elevation angle satellites can be corrected and used, as opposed to the cases where they are rejected.
   \item As shown in the static test, when the time difference of TDCP increases, its error increases as well. If the time difference is within 60 s, it can be used for graph optimization as a loop closure with good accuracy.
\end{itemize}

The experimental environment considered in this study was an open-sky environment. In this situation, the use of a satellite with a low elevation angle is expected to reduce the dilution of precision and improve the accuracy, especially in the vertical direction. However, in urban environments, low-elevation satellites are likely to be non-line-of-sight (NLOS) satellites, and a large elevation mask is required to eliminate NLOS satellites, resulting in a trade-off. When this method is applied to urban environments where NLOS multipath occur, the process for determining NLOS multipath is considered essential. The application of this method to NLOS multipath environments will be a future challenge. 

\section{Conclusion}
In this paper, we propose a highly accurate trajectory estimation method for a mobile robot using GNSS TDCP observations. Additionally, we propose a graph structure in which the amount of cycle slip is included in the TDCP factor. The nodes of the number of cycle slips are added together with each carrier phase measurement, and the constraint of the time variation of the cycle slip is added to the graph. By adjusting the variance of the time variation of cycle slip based on the possibility of its occurrence, it was shown that cycle slip can be properly detected, estimated, and corrected by graph optimization. In real-world static tests, the accuracy of the relative position estimation was improved using TDCP observations using low-elevation satellites. The results of the kinematic and static tests showed that the trajectory of the robot could be estimated with higher accuracy than the previously proposed method using TDCP.

\ifCLASSOPTIONcaptionsoff
  \newpage
\fi



\bibliographystyle{IEEEtran}
\bibliography{IEEEabrv,RAL_2022}

\end{document}